\documentclass[conference]{IEEEtran}
\IEEEoverridecommandlockouts
\def\IEEEtitletopspace{18pt}
\usepackage{cite}
\usepackage{amsmath,amssymb,amsfonts}
\usepackage{algorithmic}
\usepackage{graphicx}
\usepackage{flushend}
\usepackage[capitalise]{cleveref}
\usepackage{textcomp}
\usepackage{xcolor,colortbl}
\usepackage[top=54pt, left=54pt, right=54pt, bottom=54pt]{geometry}% 
\def\BibTeX{{\rm B\kern-.05em{\sc i\kern-.025em b}\kern-.08em
    T\kern-.1667em\lower.2ex\hbox{E}\kern-.125emX}}
\begin{document}

\definecolor{Gray}{gray}{0.75}
\definecolor{Grey}{gray}{0.92}

\newcolumntype{a}{>{\columncolor{Grey}}c}

\title{\LARGE \bf
Aerial Tensile Perching and Disentangling Mechanism for Long-Term Environmental Monitoring}
\author{ Tian Lan$^{*,1}$, Luca Romanello$^{*,1,2}$, Mirko Kovac$^{2,3}$, Sophie F. Armanini$^{1}$, Basaran Bahadir Kocer$^{2,4}$ \\
\thanks{
© 2024 IEEE.  Personal use of this material is permitted.  Permission from IEEE must be obtained for all other uses, in any current or future media, including reprinting/republishing this material for advertising or promotional purposes, creating new collective works, for resale or redistribution to servers or lists, or reuse of any copyrighted component of this work in other works.
\par $^{*}$contributed equally
\par $^{1}$eAviation Laboratory, School of Engineering and Design, TU Munich.
\par $^{2}$Aerial Robotics Laboratory, Imperial College London.
\par $^{3}$Laboratory of Sustainability Robotics, EMPA, Dübendorf, Switzerland.
\par $^{4}$School of Civil, Aerospace and Design Engineering, University of Bristol.}
}

\maketitle

\begin{abstract}
Aerial robots show significant potential for forest canopy research and environmental monitoring by providing data collection capabilities at high spatial and temporal resolutions. However, limited flight endurance hinders their application. Inspired by natural perching behaviours, we propose a multi-modal aerial robot system that integrates tensile perching for energy conservation and a suspended actuated pod for data collection. The system consists of a quadrotor drone, a slewing ring mechanism allowing 360° tether rotation, and a streamlined pod with two ducted propellers connected via a tether. Winding and unwinding the tether allows the pod to move within the canopy, and activating the propellers allows the tether to be wrapped around branches for perching or disentangling. We experimentally determined the minimum counterweights required for stable perching under various conditions. Building on this, we devised and evaluated multiple perching and disentangling strategies. Comparisons of perching and disentangling manoeuvres demonstrate energy savings that could be further maximized with the use of the pod or tether winding. These approaches can reduce energy consumption to only 22\% and 1.5\%, respectively, compared to a drone disentangling manoeuvre. We also calculated the minimum idle time required by the proposed system after the system perching and motor shut down to save energy on a mission, which is 48.9\% of the operating time. Overall, the integrated system expands the operational capabilities and enhances the energy efficiency of aerial robots for long-term monitoring tasks.

%% will be updated!

% We also provided the minimum required idle time after the system perches and shuts down the motors to exploit the energy-saving feature for a mission, which is 48.9\% of operational time.

%Our proposed robotic system is energy efficient if it spends at least 48.9\% of its operation time in a perched state.

%% Background
%% Problems
%% Solutions
%% Result of this work

\end{abstract}

% \begin{IEEEkeywords}
% component, formatting, style, styling, insert
% \end{IEEEkeywords}

\section{Introduction}

\subsection{Canopy Access with Aerial Robots}

The forest canopy, which comprises the collective foliage, twigs, and branches of a forest, plays a vital role in determining species richness, atmospheric exchange, and understory microclimate \cite{parker1995structure}. However, studying forest canopies poses significant challenges due to their physical location (20 to 60 meters above the ground) and complex three-dimensional environments that are often dense and GPS-denied \cite{lowman2021life}. While scientists can observe forests from above using high-resolution optics on satellites and aeroplanes, dense canopies obstruct views of the forest floor and the interior of the canopy. Consequently, the primary method of obtaining data from inside the canopy still relies on manual intervention, such as using ropes, towers and walkways, ladders, and cranes to place the instruments in the field \cite{kocer2021forest}. These methods to access canopies are laborious, expensive, and pose safety risks \cite{de2021forest, barker2001forest}. As the evolution of aerial robots advances, their significant potential for environmental monitoring and exploration becomes more pronounced \cite{ho2022vision}. Sensors mounted on aerial robots offer a unique opportunity to collect data with high spatial and temporal resolution over relatively large areas, thus bridging the existing gap between field observations and traditional remote sensing by aircraft and satellites. Costs and personnel consumption are consequently significantly reduced \cite{manfreda2018use}, and the use of aerial robots within the forest canopy is explored \cite{jaakkola2017autonomous}. 
\begin{figure}[t!]
    \centering
\includegraphics[width=0.9\columnwidth]{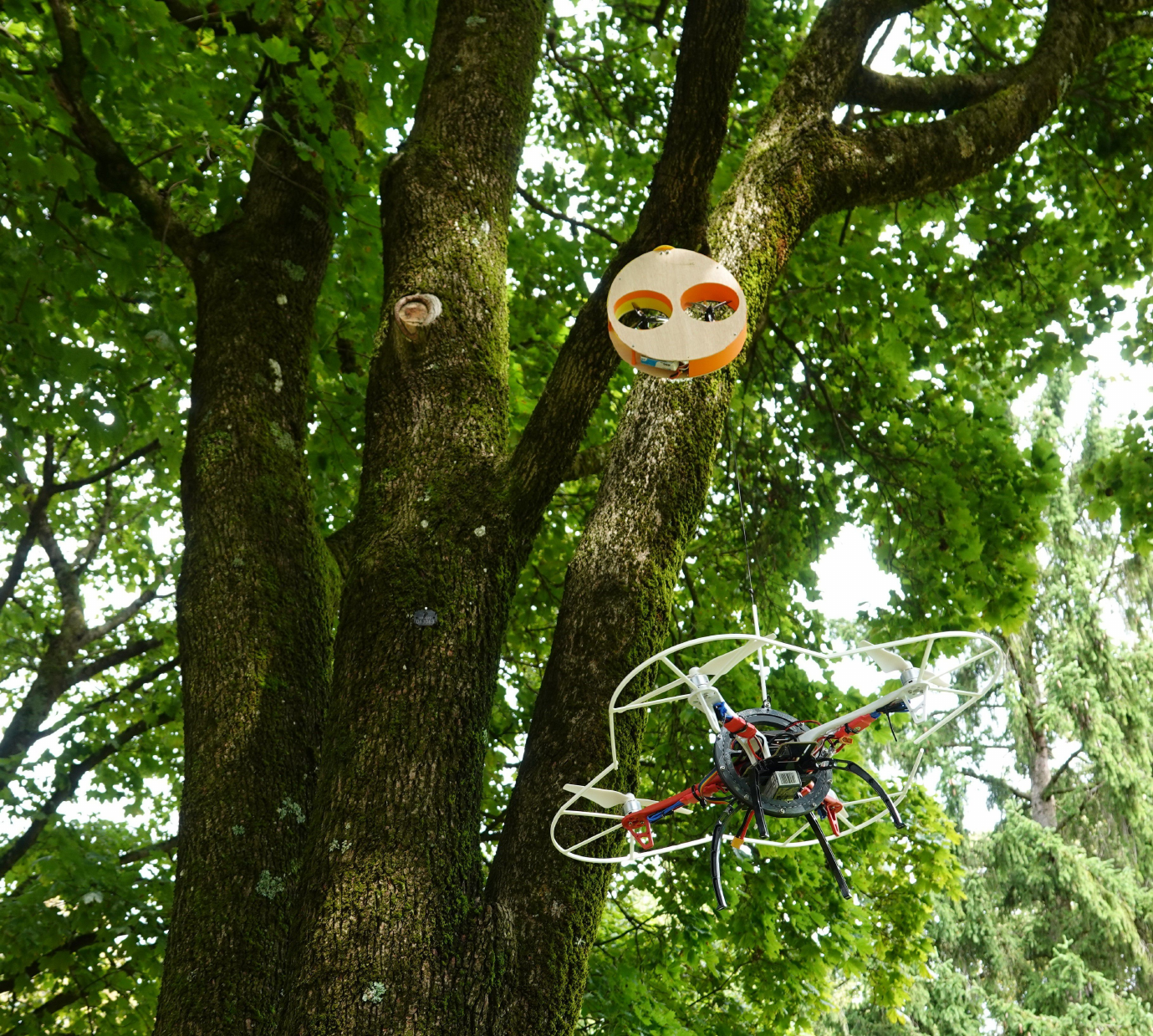}
    \caption{ Multi-modal aerial robot system under a tree canopy, able to perch and disentangle from the branch.}  \label{fig:systemIntree}
\end{figure}
\subsection{Perching Technology and Spider-inspired Pods}
% why perching

From an operational efficiency standpoint, current drone battery technology imposes significant limitations on flight time \cite{floreano2015science}. For example, DJI's medium-sized enterprise quadrotor drone offers a mere 38-minute flight time, which is shortened to approximately 20 minutes with a payload \cite{DJI2023}. This time scale hardly supports work in complex environments requiring operational complexity, including repeated landings and takeoffs to recharge the batteries \cite{kocer2019uav}. In nature, animals often perch in trees and stay there for long periods to conserve energy or to avoid danger. Perching at a height also helps them observe their surroundings. Inspired by this natural phenomenon, scientists developed the idea of drone perching, which could alleviate the massive energy expenditure incurred during long flights \cite{broers2022design}. However, ensuring a safe and reliable perching and taking off from the same position has been a challenge, especially when the perching object is unknown, and has a complex and irregular geometry (e.g. within a tree canopy). To address this challenge, various practical perching mechanisms have been proposed and customized to specific environmental conditions. \cite{roderick2021bird,broers2022design,kirchgeorg2022soft} proposed different grippers in the form of claws and microspines to perch on round objects. \cite{kirchgeorg2021hedgehog} integrates the mechanism with microspines to increase friction to hold the weight of the drone platform. \cite{nguyen2019passively} and \cite{hauf2023learning} enable the drone to be perched on tree branches through a tether. \cite{zheng2023metamorphic} developed a small, metamorphic perching aerial robot that can grasp tree branches using its multi-rotor arms. The robot's bistable arm is rigid in flight but conforms to its purpose in perching. \cite{nguyen2023soft, 10380767, askari2024crashperching} use grippers based on inflatable high-strength fabrics and bistable mechanisms to enable grasping and perching on objects of unknown shapes and sizes. Although perching reduces energy consumption overall, its efficiency varies depending on how the specific mechanism works. Among the proposed perching mechanisms, tethered perching shows excellent potential to reduce energy consumption for long-term data collection tasks due to its simplicity and stability, as well as other aspects including tree exploration from the top \cite{hauf2023learning}. The tethered perching approach also allows sensor installation after disconnecting or cutting off the tether connection. Its simplicity, energy efficiency, and lightweight design make it adaptable and environmentally friendly in a way that reduces noise. Furthermore, this simple configuration is compatible with additional mechanisms or sensors, showing its potential for integration and versatility. However, as stated in \cite{hauf2023learning}, the upside-down position of the drone after perching has consistently been an issue. This inconvenient take-off position makes disentangling after perching more complex and less energy-efficient. Another innovative way to enhance the versatility of aerial robots is the use of suspended pods. When mounted on drones, these pods offer a promising solution for long-duration missions. Drone operations may impact wildlife due to high-frequency propeller noise. Furthermore, the need for ever-increasing data volumes still poses problems for methods that only use drones directly to collect data \cite{francis2013framework,brouvcek2014effect,pater2009recommendations}. To address these issues, a solution is to switch off the motors and apply tethered perching combined with a suspended pod. This method mitigates potential disturbances to the ecological balance while exploiting the individual strengths of the drone and the suspended pod. \cite{mckerrow2007design} investigated the design and modelling of a robot that swings at the end of a tether. However, this research did not further investigate specific use scenarios for tethered robots. Later, \cite{polzin2023heading} presented a tethered bi-copter with horizontal propellers to aid in the environmental monitoring of glaciers. Recently, \cite{zhang2017spidermav} has taken spiders as inspiration and \cite{kirchgeorg2022multimodal,kirchgeorg2023design} proposed the design of a tethered, suspended pod for tree canopy research. The pod has two horizontal propellers and is suspended from a tether. 

    \begin{figure}[t!] %figure************
    \centering
    \includegraphics[width=0.9\columnwidth]{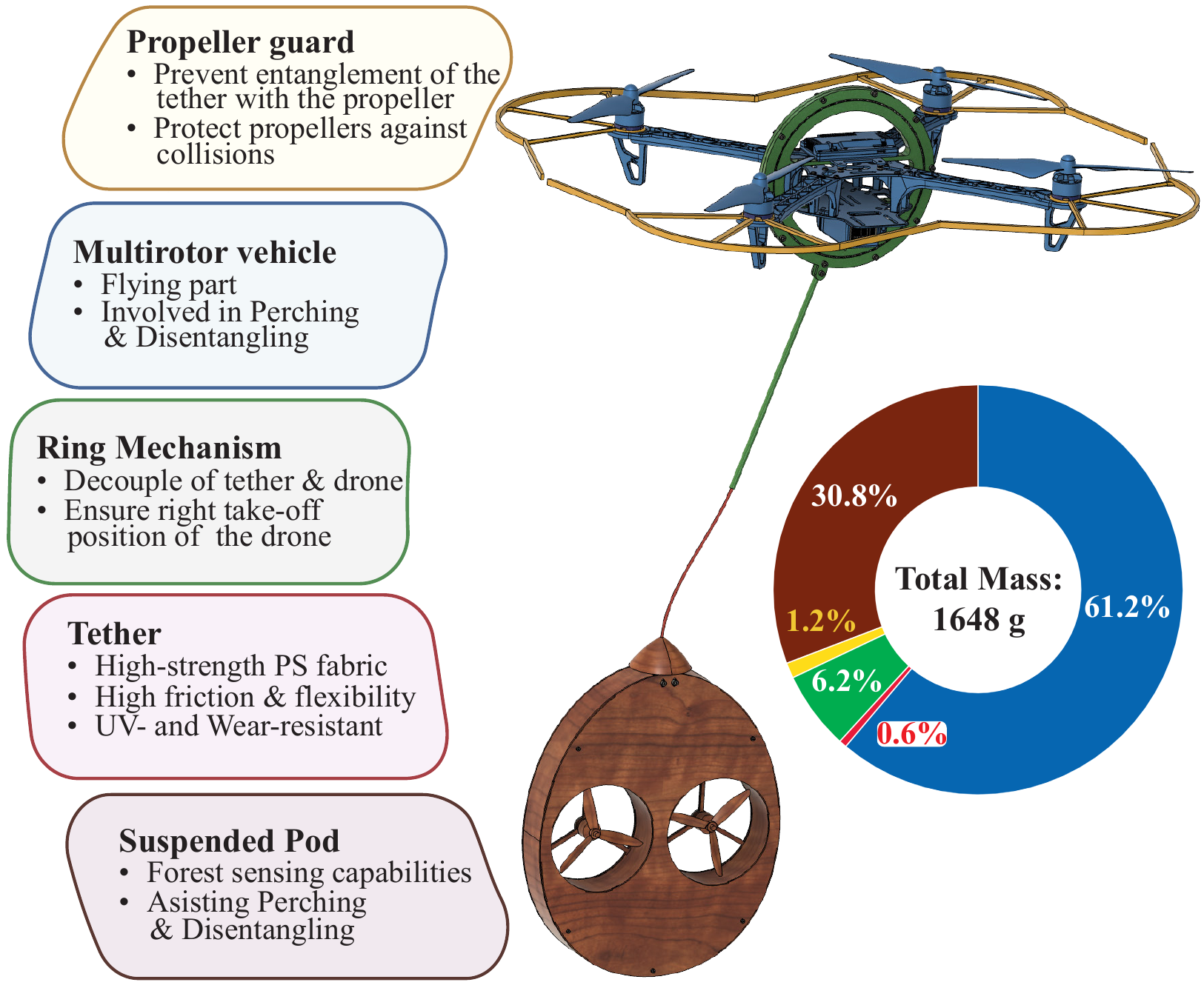}
    %{Figures/pod_drone_design_biggerText_only Slewbearing.pdf}
    \caption{Illustration of the multi-modal robot system design and its weight distribution.}
    \label{fig:complete_design}
    \end{figure}

\subsection{Tethered Perching for Data Collection}

This work demonstrates a multi-modal aerial robot system, as shown in Fig.~\ref{fig:systemIntree}. By combining a perching device and a suspended pod, the aerial robot gains the ability to perch for energy conservation and can deploy specialized equipment or sensors from the pod to collect data in complex environments such as within the tree canopy, reducing the workload of ecologists and increasing their safety. This integrated approach maximizes the operational potential of the aerial robot and allows it to adapt to a wide range of environmental monitoring scenarios. The system consists of five main components (Fig. \ref{fig:complete_design}): a drone, propeller guards, a ring mechanism, a tether, and a suspended pod. Thanks to the propeller and tether winding system on the suspension pod, it can not only manoeuvre within the canopy and collect data, as shown in  \cite{kirchgeorg2022multimodal,kirchgeorg2023design}, but also assist with perching and decoupling to further improve energy efficiency. The main contributions of this work are summarized as follows:

\begin{enumerate}
\item Introducing a multi-modal aerial robot system for environmental data collection, that integrates a drone and a flexible tethered pod, allowing the system to perch on a tree branch and disentangle.
\item Designing a ring mechanism to ensure that the aerial system can take off from an upside-down position by removing the dependency of a fixed point anchoring from the drone body. 
\item Experimentally evaluating the effectiveness of different perching and disentangling strategies implemented through drone body, proposed pod and tether systems. %that coul lead ecologist adopting or methodology (or autonomous way of environmental sensing)
\end{enumerate}

%To the best of our knowledge, this is the first study combining the tethered perching and disentangling strategy with the tethered robot. 
\section{System Design}

As shown in Fig. \ref{fig:complete_design}, the multi-modal aerial robot system is composed of a drone with a mounted ring mechanism and a suspended pod with two propellers. The pod and the ring mechanism on the drone are connected by a tether. 

\subsection{Drone and Ring Mechanism}
\begin{table}[b!]

\caption{Specifications of the Drone and Pod.}
\begin{center}
\scalebox{0.97}{
\begin{tabular}{|l|l|l|}
\hline
\rowcolor{Gray}
{\textbf{Components}} & {{\textbf{Drone}}} & {\textbf{Pod}} \\
\hline
{Frame} & {Standard F450} & {Custom} \\
\hline
{Flight controller}&{Ardupilot APM2.8} &{Pixracer R15} \\
\hline
{Propeller Motor} & {DJI 2212 920KV} & {Flash hobby A1408 M5}\\
\hline
{Retraction servo motor} & {} & {SPT5325LV-360} \\
\hline
{ESC} &  {Simonk 30A} &  {EMAX bullet pro 35A} \\
\hline
{Propeller} &  {9450} & {4043} \\
\hline
{Battery} &  {4S 1300 mAH Lipo} &  {4S 450 mAH Lipo}\\
\hline
{Total weight} &  {1008 g} &  {508 g} \\
\hline
\end{tabular}}
\label{tab: components}
\end{center}
\end{table}

 \begin{figure}[t!] 
    \centering
   \includegraphics[width=1\columnwidth]{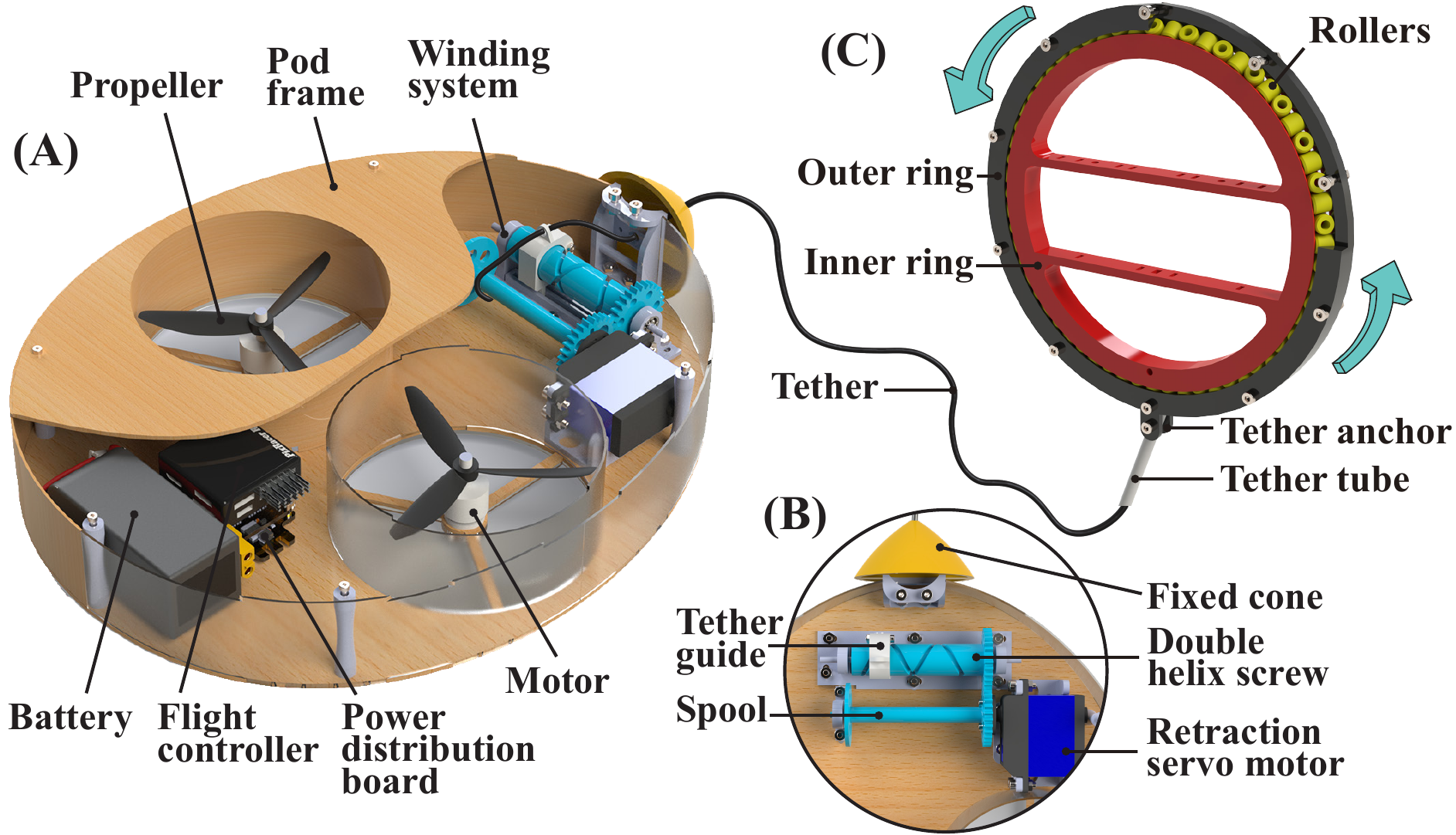}
    \caption{ Illustration of the ring mechanism and pod design: (A) 3D model of the pod. (B) Tether winding system. (C) 3D model of the ring mechanism.  }
    \label{fig:componentsdesign}
    \end{figure} 
We assembled a drone according to the specifications in Table \ref{tab: components} and installed a ring mechanism on the drone as shown in Fig. \ref{fig:componentsdesign} (C). This mechanism serves a dual purpose. Firstly, it enables the tether to rotate freely around the drone, ensuring that the drone's propellers maintain an upward orientation after perching. Secondly, it mitigates issues related to a shifting centre of gravity and propeller interference, commonly observed with fixed tether anchor points. Given that the pod is suspended beneath the drone during flight, the ring mechanism is engineered to withstand both radial and axial forces. Specifically, the radial forces arise from the pod's weight, while the axial forces are generated by the pod's swinging motion. To meet these requirements, we used a slewing bearing structure. %This specialized type of rolling-element bearing allows for both rotational and axial movement between components.

The ring mechanism features an outer and an inner ring, separated by rolling elements. This configuration enables it to accommodate axial, radial, and moment loads, making it particularly well-suited for applications that demand smooth, controlled rotation. As depicted in Fig. \ref{fig:complete_design}, the inner ring is affixed to the drone frame, whereas the outer ring is linked to the tether. This design ensures that the tether can rotate freely, independent of the drone's movements after the system perches on a branch. To achieve a lightweight construction while maintaining functionality, the slewing ring incorporates a single row of 45° crossed rollers, as described in \cite{kania2012catalogue}. The total weight of this mechanism is 102 grams. To further optimize the system, a tether tube has been installed at the tether anchor point. This feature serves to constrain the tether's movement within the propeller guard, effectively preventing it from coming into contact with the propellers.

 \subsection{Suspended Pod}
 
%     \begin{figure}[t!] 
%     \centering
%     \includegraphics[width=0.8\columnwidth]{Figures/pod design_v2.pdf}
%     \caption{ Illustration of the pod design. (A) 3D model of the pod. (B) Fixed cone mounted on top of the pod. (C) Tether winding system. 
%     }
%     \label{fig:Poddesign}
%     \end{figure} 
% \subsubsection{ Pod Design }  

The pod is actuated by a retractable tether and a set of ducted propellers. This multifunctional design enables the pod to assist in the perching operation, navigate through forest canopies, and collect data via onboard sensors. The design details of the pod are illustrated in Fig. \ref{fig:componentsdesign} (A), and its detailed specifications are available in Table \ref{tab: components}.

The frame of the pod is constructed from 2 mm plywood and serves as the foundational element for all other components. The egg-shaped design, as depicted in Fig. \ref{fig:componentsdesign} (A), provides multiple structural benefits. Specifically, the egg's curvature contributes to a uniform distribution of external forces, as well as superior resistance to deformation, as substantiated by \cite{solomon2010eggshell}. Additionally, the shape - narrow at both ends and wider in the middle - minimizes the risk of entanglement when navigating through forest canopies. Fig. \ref{fig:componentsdesign} (B) shows the cone positioned atop the pod. This cone serves a dual purpose: it acts as a protective shield against external debris and water infiltration - a crucial feature given the pod's operation in densely wooded areas - and functions as a buffer between branches and the pod's structure.

%esigned in an egg-like shape, as depicted in Fig\ref{fig:componentsdesign}, this form factor offers several structural benefits.
\subsubsection{Winding System}

Experimental observations revealed that without controlled winding, the tether tends to accumulate in the spool's centre. This results in suboptimal use of the available spool length and creates a bulky spool that can interfere with the pod's limited internal space, compromising the retraction functionality. To address this issue, we engineered a winding system that incorporates a spool, a servo motor, a double helix screw, and a tether guide, as illustrated in Fig. \ref{fig:componentsdesign} (B). The tether is secured to the spool after threading it through the fixed cone and the tether guide. When activated, the servo motor initiates spool rotation, which in turn drives the double helix screw via a gear mechanism. This action generates a lateral reciprocating motion in the tether guide, ensuring uniform tether alignment on the spool. Except for the bearings, all winding system components are fabricated using fused deposition manufacturing with polylactic acid (PLA).

\subsubsection{Minimum Counterweight}

In our design evaluations, the primary objective is to minimize the weight of the suspended pod to reduce energy consumption during flight. The perching and disentangling processes involve scenarios with either one or two wrapping loops of the tether. If the drone's minimum counterweight requirements are not met, the tether will lack sufficient friction to maintain a perched position. Consequently, assessing the minimum counterweight in these scenarios is crucial for the success of the perching process. Test variables include the number of tether loops wrapped around the branch (ranging from 1 to 3) and the angle of the branch relative to the ground (varying from 0° to 30°). The tether used in these tests has a 1 mm diameter and is constructed from polyester fabric. The test branch features a smooth surface and a diameter of 73.2 mm. To clarify the relationships between these variables, all measured counterweights were normalized by dividing them by the drone's take-off weight.

\begin{figure}[t!]
    \centering
 \includegraphics[width=1\columnwidth]{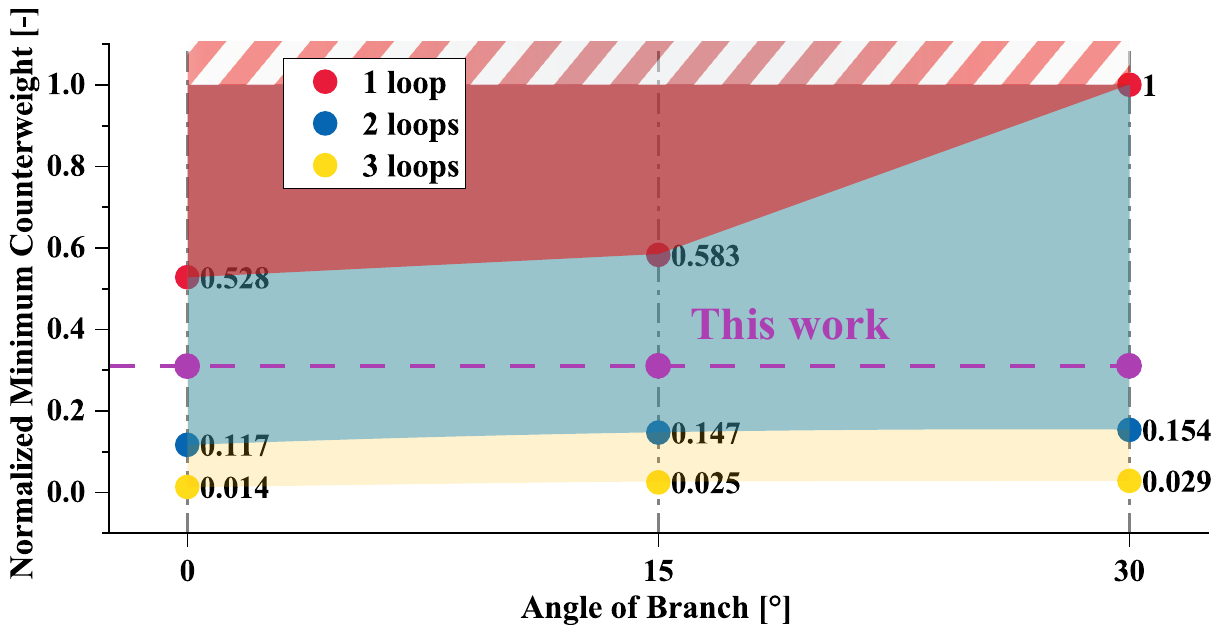}
    \caption{ Minimum counterweights for different wrapping loops and branch angles. Feasible areas are indicated by colours for each different wrapping loop.}
    \label{fig:pod_radar}
    \end{figure}

Fig. \ref{fig:pod_radar} displays the minimum counterweights required for perching. Notably, the tether becomes unstable when the branch angle is at 30° with only a single loop around it. This scenario is represented by a weight ratio of 1. The figure aids in identifying the viable payload weight for a tethered system. Here, "viable" implies that the tether can produce enough friction to maintain stability on the branch, or in other terms, that the weight ratio exceeds its minimum threshold. These viable regions are highlighted with corresponding colours. It is crucial to note that the weight ratio should never exceed 1, as this would imply the payload outweighs the drone itself. In the present system, the pod's weight accounts for 30.8\% of the total weight. When using only one loop for wrapping, this weight falls below all identified minimum counterweights, indicating that at least two loops are necessary for successful perching. Thus, we subsequently utilized two wrapping loops to validate the proposed system. Additionally, our observations reveal that the required counterweight decreases with an increasing number of wrapping loops, but rises as the branch angle increases. Comparative analysis of the minimum counterweights for various scenarios highlights the significant impact of the number of wrapping loops on tether friction. By contrast, the effect of the branch angle appears to be minimal for the cases that require more than one loop. This can be attributed to the fact that the increase in contact surface, achieved by adding more loops, has a far greater influence on friction than other factors such as branch angle. This phenomenon is accurately described by the Capstan equation and a more comprehensive explanation of this effect is detailed in \cite{lim2022design}. 

\section{Working Principles} 

    \begin{figure*}[t!] %figure************
    \centering
    \includegraphics[width=0.81\textwidth]{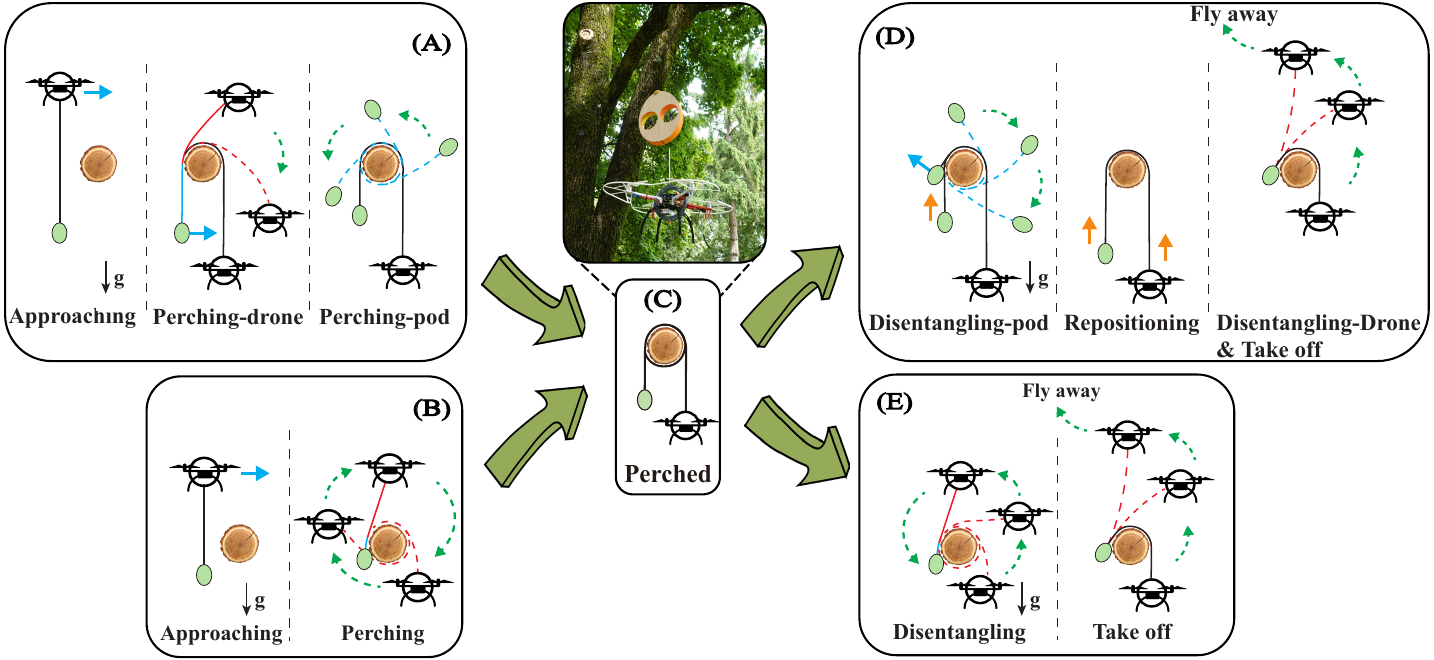}
    
    \caption{Schematic of strategies for (A) duo-perching (B) solo-perching (D) duo-disentangling (E) solo-disentangling, where (C) is the perched state.}
    \label{fig:Perchingstrategies}
    \end{figure*}
    
By employing the proposed system, we aim for sustained periods of data collection with aerial robots which can perch on tree branches and shut down the motors. 
\subsection{Perching} 
%The perching strategy can vary depending on the components involved: it may utilize either the pod and drone together or just the drone alone. To distinguish between these approaches, we refer to the use of both the pod and drone as "duo-perching," and the use of only the drone as "solo-perching." 
The perching strategy can differ based on the components utilized: it can involve both the pod and drone (referred to as "duo-perching") or only the drone (referred to as "solo-perching"). A schematic illustration of these strategies is presented in Fig. \ref{fig:Perchingstrategies}. The perching method relies on the friction between the tether and the branch surface, augmented by the self-weight of the pod, to counterbalance the drone's weight. Duo-perching is executed in a three-step process, as depicted in Fig. \ref{fig:Perchingstrategies} (A). Initially, the system approaches the target branch while the pod's winding system releases an adequate length of tether. As previously analyzed, our target is to achieve two wrapping loops. The first loop is performed by the drone, which flies over the branch and descends below the pod. Once the drone stabilizes, the pod activates its propellers to execute the second loop. Utilizing a blend of thrust and centrifugal force, the pod manoeuvres itself around the branch. Fig. \ref{fig:Perchingstrategies} (B) depicts the solo-perching scenario. In this case, the drone is responsible for completing both wrapping loops around the branch, effectively carrying out the entire perching process on its own.

\subsection{Disentangling} 

%As delineated in Fig. \ref{fig:Perchingstrategies}, we have implemented both duo-disentangling and solo-disentangling strategies. 
We also implemented "duo-disentangling" and "solo-disentangling" strategies. The duo-disentangling process consists of three steps, as illustrated in Fig. \ref{fig:Perchingstrategies} (D). Initially, the pod is retracted until it is in close proximity to the branch, thereby minimizing the radius needed for disentangling. Subsequently, the pod reverses the first loop, effectively mirroring the perching process but in the opposite direction. Manoeuvre around the branch can be achieved by retracting the tether or using the pod's propellers. Once this step is complete, the tether is further retracted in preparation for the drone's takeoff. It is crucial at this stage to maintain a minimal distance between the pod and the drone to prevent any pendulum effects and ensure a safe takeoff. The disentangling of the second loop occurs automatically as the drone departs from the branch. In the case of solo-disentangling, depicted in Fig. \ref{fig:Perchingstrategies} (E), the drone is solely responsible for the entire disentangling procedure. It accomplishes this by executing two loops around the branch, reversing the perching steps. Concurrently, the pod is retracted to shorten the length of the tether, enhancing stability and ensuring a safer drone takeoff.

\section{Tests and Results}

A set of analyses and tests were conducted to both validate the robustness of the system's design and assess its potential for energy conservation. The flight tests were conducted on a fixed branch structure simulated forest canopy conditions.

\subsection{Energy Consumption}
\begin{figure}[b!]
    \centering
        \includegraphics[width=0.9\columnwidth]{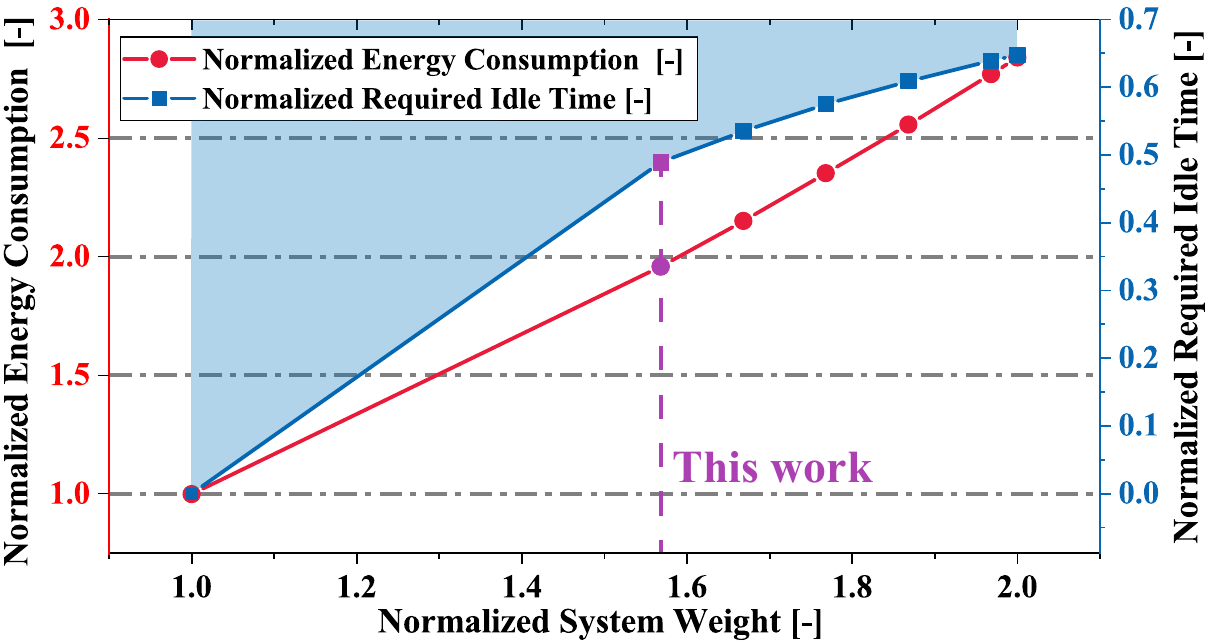}
    \caption{ Normalized energy consumption and required idle time over system weight.}
    \label{fig:idletime}
    \end{figure}
As evidenced in Fig. \ref{fig:idletime}, our analysis indicates that tensile perching can yield substantial energy savings, but it requires a minimum perching duration. Specifically, we determined that a minimum of 48.9\% of the total operational time must be spent in a perched state to gain any energy-saving benefits for our proposed robotic system. Moreover, we have identified the minimum idle times correlated with increments of system weights that are necessary to achieve these energy savings, which are applicable to various design cases regardless of the specific platform. Fig. \ref{fig:idletime} further shows that any idle time exceeding the specified values, adjusted for various pod weights, will result in energy conservation. Our tethered perching system offers a viable solution for drones to minimize energy expenditure during extended aerial data collection missions.
\begin{figure*}[t!]
\centering
    \includegraphics[width=0.9\textwidth]{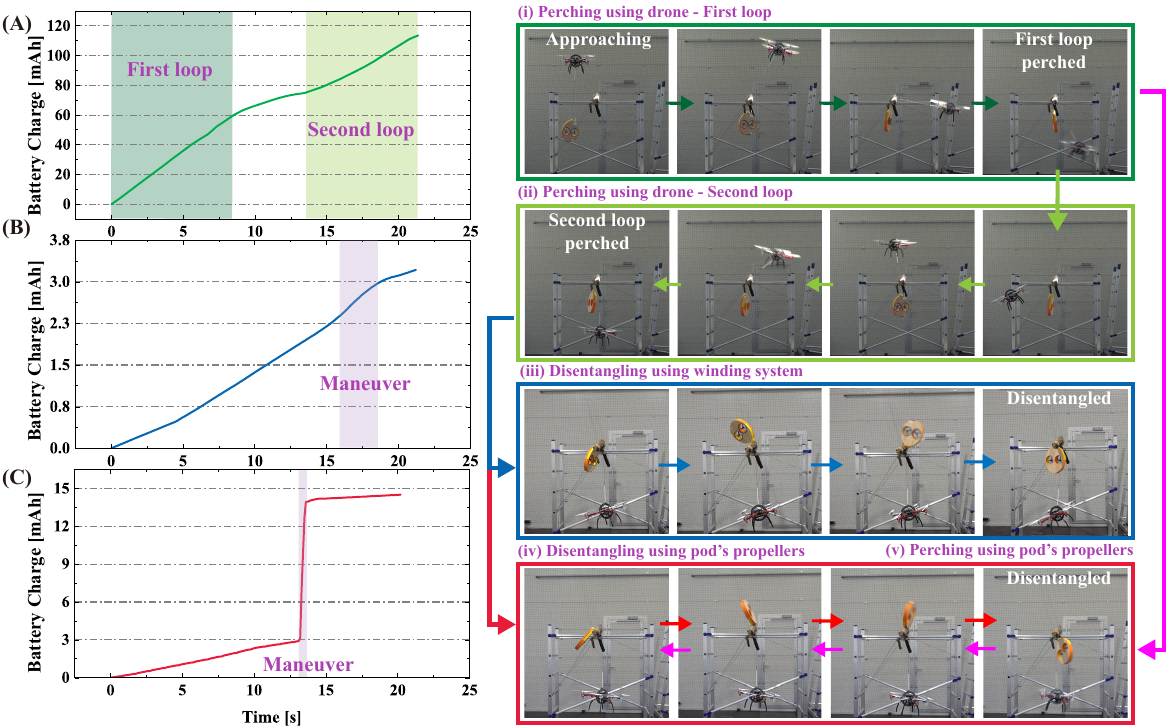}
\caption{ Battery charge for (A) perching using the drone, (B) disentangling using the winding system, and (C) disentangling using the pod's propellers. The experimental snapshots of the aerial robot system performing perching and disentangling are shown in (i)-(v). }
\label{fig:TotalCurrent}
\end{figure*}
%Our system involves the drone with a specialized mechanism comprising a ring, tether, and pod, enabling it to securely perch on stable structures such as tree branches or power lines while engaged in data collection. 

Subsequently, we carried out flight tests to evaluate the energy efficiency of various perching and disentangling strategies. Given the reciprocal nature of perching and disentangling manoeuvres, we operated under the assumption that both processes would consume equivalent amounts of energy, thereby eliminating the need for duplicate data collection. The energy expenditure for both flight and perching activities is determined by battery charge data, which was accurately measured using the drone's onboard flight controller. Fig. \ref{fig:TotalCurrent} illustrates the battery charge required and experimental snapshots for drone-based perching and pod-based disentangling. The drone-based perching can be divided into two parts: the first loop and the second loop. As explained earlier, the drone completes two wrapping loops around the branch for a solo-perching strategy (Fig. \ref{fig:TotalCurrent} (i) and (ii)). For the duo-perching strategy, the first loop is achieved by the drone and the second loop by the pod using onboard propellers (Fig. \ref{fig:TotalCurrent} (i) and (v)). Fig. \ref{fig:TotalCurrent} (A) shows the energy usage for the first and second loops. The drone hovered briefly between the two to stabilize itself. The first loop consumes about 60 mAh of power, while the second loop requires about 49 mAh. Overall, their energy consumption is comparable, but the first loop consumes more energy. This outcome is consistent with our expectations because the drone requires extra time for adjustments, including tether length and distance to the branch before perching the first loop. 
For pod-based maneuvers, we explored two methods for comparison: disentangling through tether retraction, facilitated by the onboard winding system, and disentangling via propeller-generated thrust. In the first method, the unique shape of the pod and the fixed cone at its apex enable automatic movement around the branch as the tether is retracted (Fig. \ref{fig:TotalCurrent} (iii)). In the second method, the propellers generate a brief burst of thrust, allowing the pod to swing around the branch by inertia (Fig. \ref{fig:TotalCurrent} (iv)). This is possible because friction between the tether and the branch prevents the free sliding of the pod. Fig. \ref{fig:TotalCurrent} (B) and (C) indicate a notable difference in energy consumption between these two methods. Specifically, the winding system consumes 0.73 mAh, while the propeller-based method requires 11 mAh$-$roughly 15 times more energy. It is worth noting, however, that disentangling using the winding system demands a longer time and higher control accuracy. The use of the pod and the winding system results in significantly lower energy consumption, only 22\% and 1.5\%, respectively, of the energy expenditure to disentangle purely with the drone (equivalent to perching the second loop). As part of an auxiliary investigation, the method's repeatability was assessed. Twenty-five iterations of drone perching and disentangling were conducted, resulting in successful completion in 23 instances. In relation to the pod, it successfully disentangled on every occasion utilizing both propellers and the winding system. Additionally, it achieved perching in 9 out of 10 attempts, with the weak yaw control being the limiting factor.

\subsection{Critical Distances}

\begin{figure}[b!]
    \centering
        \includegraphics[width=1\columnwidth]{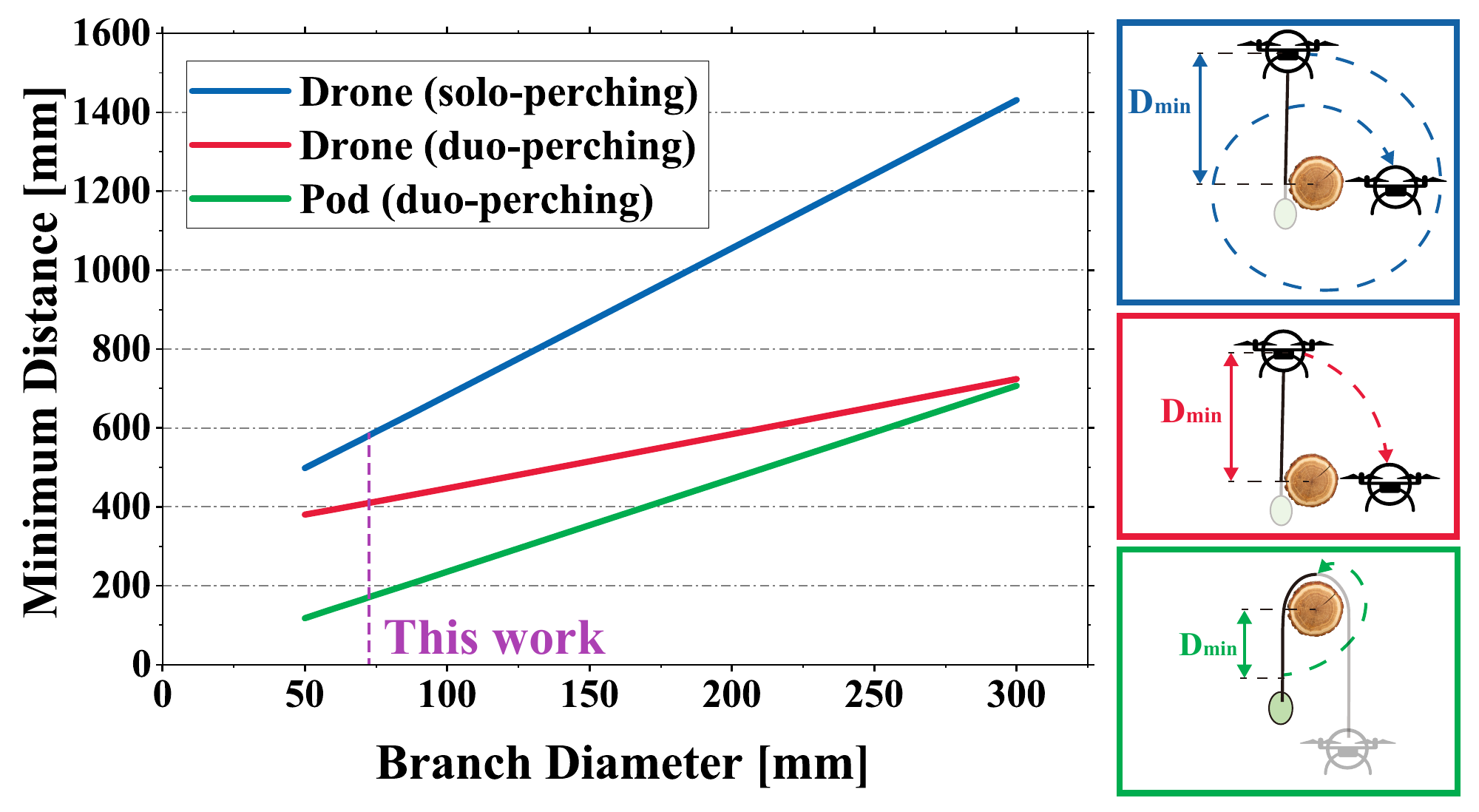}
    \caption{ Critical distances for perching with the pod or the drone with increasing diameters of the branch. }
    \label{fig:DistancePod}
\end{figure}

A critical consideration when using the pod's propellers for perching is the minimum distance between the pod and the branch. If this distance drops below $D_{ \rm min}$, the perching manoeuvre may fail as the pod may be unable to navigate around the branch. Similarly, the distance between the drone and the branch is a vital parameter for identifying suitable perching locations. To better understand these minimum distance requirements, we have carried out calculations for various branch diameters, with results presented in Fig. \ref{fig:DistancePod}. These values represent the minimum distances necessary for successful perching, thus serving as benchmarks for different perching strategies.

%A key factor to consider when using pod's propellers to perch is the minimum distance between the pod and the branch. If this distance falls below $D_{ \rm min}$, the perching maneuver may fail as the pod may be unable to navigate around the branch. Similarly, the distance between the drone and the branch is a vital parameter for identifying suitable perching locations. The required free space around the perching object can be determined based on this. To better understand these minimum distance requirements, we have carried out calculations for various branch diameters. The results are presented in Fig. \ref{fig:DistancePod}. Specifically, we determined the minimum distances correlated with increments of branch diameters that are necessary to achieve successful perching. In other words, the distances for different perching strategies must exceed these specified values to ensure successful perching.

\section{Conclusion}
In this work, we introduced a multi-modal aerial robot system designed for long-duration environmental monitoring in complex terrains like forest canopies. Utilizing a tensile perching mechanism and a specialized pod, the system achieved significant energy savings, requiring perching and staying idle for 48.9\% of operational time to realize these energy savings. Various perching and disentangling strategies were tested and optimized, highlighting the significant benefits of the tether winding system, which is up to 15 times more energy-efficient than propeller-based methods. Critical distances for successful perching were also presented, contributing valuable insights for future missions. These results considerably enhance the capabilities and energy efficiency of aerial robots for long-term monitoring tasks while minimizing environmental disturbance, with important implications for ecological research. Future work will focus on miniaturizing system components, incorporating additional sensing capabilities, and advancing the control and autonomy of the system through more extensive field trials in real-world environments.

\section*{Acknowledgment}
The authors would like to acknowledge and thank Diego Pérez González for his help in testing the system.

%\section*{References}

 \bibliographystyle{IEEEtran}
 \bibliography{mybib}

% \begin{thebibliography}{}

% \end{thebibliography}
\vspace{12pt}
\color{red}

\end{document}